\documentclass[10pt,twocolumn,letterpaper]{article}

\usepackage{cvpr}              
\definecolor{cvprblue}{rgb}{0.21,0.49,0.74}
\usepackage[pagebackref,breaklinks,colorlinks,allcolors=cvprblue]{hyperref}

\usepackage{graphicx}
\usepackage{amsmath}
\usepackage{amssymb}
\usepackage[utf8]{inputenc} 
\usepackage[T1]{fontenc}    
\usepackage{url}            
\usepackage{booktabs}       
\usepackage{amsfonts}       
\usepackage{nicefrac}       
\usepackage{microtype}      
\usepackage[table,dvipsnames]{xcolor}         
\usepackage{colortbl}
\usepackage{amsmath}
\usepackage{makecell}
\usepackage{multirow}
\usepackage{float}
\usepackage{pifont}
\usepackage{arydshln} 

\usepackage[accsupp]{axessibility}
\usepackage{times}
\usepackage{epsfig}
\usepackage{graphicx}
\usepackage{amsmath}
\usepackage{amssymb}
\usepackage{colortbl}
\usepackage{booktabs} 
\usepackage{multirow}
\usepackage{makecell}
\usepackage{float}
\usepackage{enumitem} 
\usepackage{arydshln}
\usepackage{bbding}
\usepackage{animate}
\usepackage{grffile}
\usepackage{siunitx} %
\usepackage{tabularx}
\usepackage{bbm}
\usepackage{pifont}     %
\usepackage{amsfonts}
\usepackage{tikz}
\usepackage{pifont}
\newcommand{\project}{WildPose} 

\colorlet{colorFst}{Green!25}       
\colorlet{colorSnd}{SpringGreen!45} 
\colorlet{colorTrd}{Yellow!30}      
\colorlet{colorLow}{darkgray!30}    
\newcommand{\fs}{\cellcolor{colorFst}\bf}   
\newcommand{\nd}{\cellcolor{colorSnd}}      
\newcommand{\rd}{\cellcolor{colorTrd}}      

\def\paperID{14407} 
\def\confName{CVPR}
\def\confYear{2026}

\title{WildPose: A Unified Framework for Robust Pose Estimation in the Wild}

\author{
Jianhao Zheng$^{1}$ \quad 
Liyuan Zhu$^{1}$ \quad 
Zihan Zhu$^{2}$ \quad 
Iro Armeni$^{1}$ \\
$^{1}$Stanford University\qquad
$^{2}$ETH Z\"urich \qquad
\\
\href{https://wildpose.github.io/}{wildpose.github.io}
}

\newcommand{\bomega}{\boldsymbol{\omega}}

\newcommand{\figref}[1]{Fig.~\ref{#1}}
\newcommand{\secref}[1]{Sec.~\ref{#1}}

\newcommand{\eqnref}[1]{Eq.~\eqref{#1}}
\newcommand{\tabref}[1]{Table~\ref{#1}}

\makeatletter
\DeclareRobustCommand\onedot{\futurelet\@let@token\@onedot}
\def\@onedot{\ifx\@let@token.\else.\null\fi\xspace}

\def\wrt{wrt\onedot}

\makeatother

\newcommand{\boldparagraph}[1]{\vspace{0.2em}\noindent{\bf #1.}}

\renewcommand{\paragraph}[1]{\boldparagraph{#1}}

\definecolor{darkgreen}{rgb}{0,0.7,0}
\definecolor{newyellow}{rgb}{1,0.8,0.05}
\definecolor{newgreen}{rgb}{0.2,0.8,0.2}

\newcommand{\ours}{Ours\xspace}

\definecolor{lightyellow}{RGB}{255,255,180} 

\newcommand{\xmark}{\color{WildStrawberry}{\ding{55}}}%

\newcommand{\noo}{\textcolor{red}{\xmark}}
\newcommand{\yess}{\textcolor{OliveGreen}{\checkmark}}

\makeatletter
\def\adl@drawiv#1#2#3{%
        \hskip.5\tabcolsep
        \xleaders#3{#2.5\@tempdimb #1{1}#2.5\@tempdimb}%
                #2\z@ plus1fil minus1fil\relax
        \hskip.5\tabcolsep}
\newcommand{\cdashlinelr}[1]{%
  \noalign{\vskip\aboverulesep
           \global\let\@dashdrawstore\adl@draw
           \global\let\adl@draw\adl@drawiv}
  \cdashline{#1}
  \noalign{\global\let\adl@draw\@dashdrawstore
           \vskip\belowrulesep}}
\makeatother

\begin{document}
\twocolumn[{%
\renewcommand\twocolumn[1][]{#1}%
\maketitle
\vspace{-3em}
\begin{center}
    \includegraphics[width=1\textwidth]{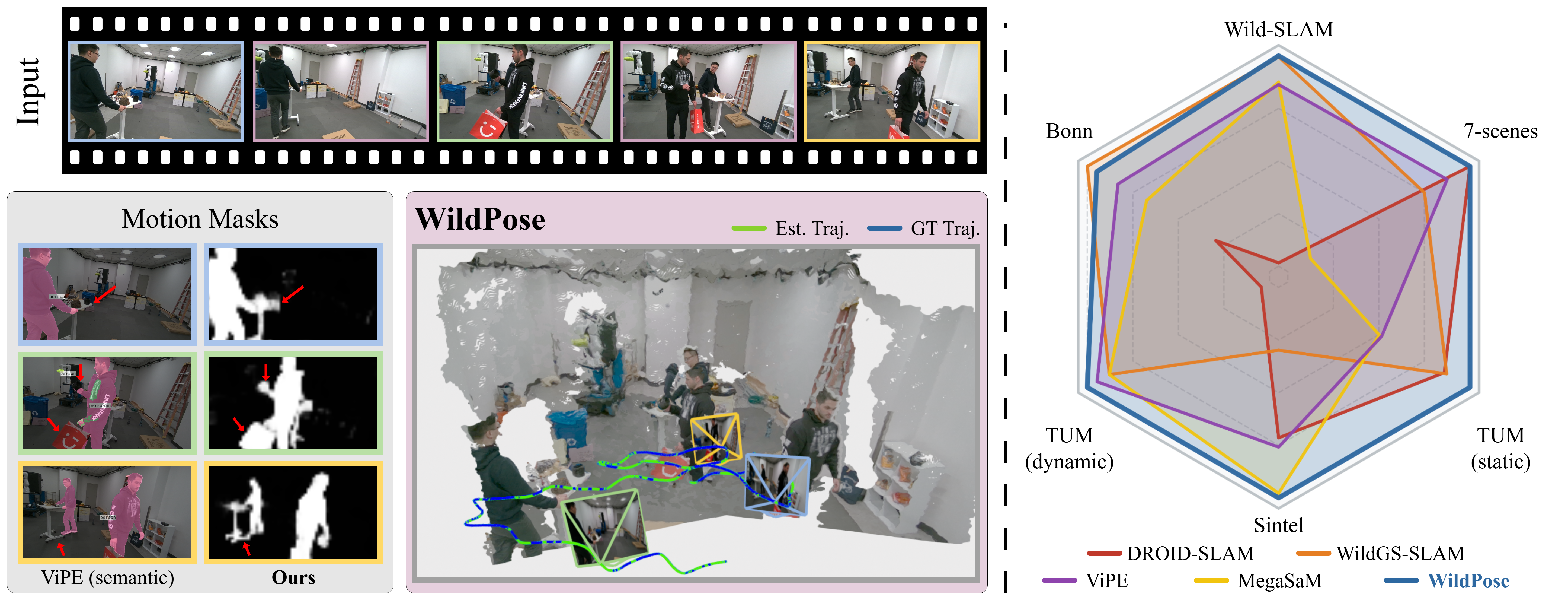}
    \vspace{-1.5em}
    \captionof{figure}{\textbf{\project{}.} \textit{Left}: Given a calibrated video sequence captured in dynamic environments, our method effectively detects the moving distractors and accurately estimates the camera trajectory, whereas methods relying on semantic segmentation fail to identify all dynamic elements.
    \textit{Right}: Pose estimation performance across multiple SLAM benchmarks, where our method achieves superior and stable performance on both dynamic and static datasets. 
    }
    \label{fig:teaser}
\end{center}
}]

\begin{abstract}
Estimating camera pose in dynamic environments is a critical challenge, as most visual SLAM and SfM methods assume inputs from static environments. While recent dynamic-aware methods exist, they are often not unified: semantic-based approaches are brittle, per-sequence optimization methods fail on short sequences, and other learned models sometimes perform badly on static-only scenes. We present \project{}, a unified monocular pose estimation framework that is robust in dynamic environments while maintaining state-of-the-art performance on static and low-ego-motion datasets. Our key insight is to connect the two powerful paradigms in modern 3D vision: the rich perceptual frontend of feed-forward models and the end-to-end optimization of differentiable bundle adjustment (BA). We achieve this by enhancing the differentiable BA pipeline in two ways. First, we introduce a new 3D-aware update operator by integrating a frozen, pre-trained MASt3R feature backbone and training the operator's subsequent layers on a diverse curriculum of static and dynamic data. Second, we propose a high-capacity motion mask detector that leverages rich, multi-level 3D-aware features from the same frozen backbone. Extensive experiments show \project{} consistently outperforms prior methods across a wide variety of benchmarks, including dynamic (Wild-SLAM, Bonn), static (TUM, 7-Scenes), and low-ego-motion (Sintel) datasets.
\end{abstract}
\vspace{-2.0em}    
\section{Introduction}
\label{sec:intro}

Estimating camera pose and 3D scene structure is a cornerstone of 3D vision, vital for applications like AR/VR and robotics. These tasks are commonly formulated as Structure from Motion (SfM) or, in the case of sequential input, Simultaneous Localization and Mapping (SLAM). While traditional geometric methods~\cite{Engel2017PAMI,Mur2015TRO,Mur2017TRO,Zhu2022CVPR,matsuki2024gaussian,zhu2024loopsplat} offered mature solutions, recent progress has been dominated by deep learning. This has led to two distinct and powerful research directions. On one hand, feed-forward SfM models~\cite{mast3r_eccv24,wang2024dust3r,wang2025vggt} are trained to directly regress 3D structure and, in some cases, camera poses. On the other hand, differentiable optimization frameworks~\cite{teed2021droid} learn an update operator to feed a differentiable bundle adjustment (BA) layer. Despite architectural differences, all these approaches largely share a critical vulnerability: they assume a static world, which causes their performance to degrade significantly in dynamic scenes.

To handle dynamic environments, a common approach is to perform motion segmentation from 2D images. Several recent methods \cite{cheng2022sg,xu2024dgslam,huang2025vipe} utilize semantic information to extract motion masks. Nevertheless, such approaches are highly sensitive to the quality of semantic segmentation and often struggle to generalize across scenes with varied and unconstrained motion patterns. Two recent works that eliminate the reliance on semantic priors are WildGS-SLAM~\cite{zheng2025wildgs} and MegaSaM~\cite{li2024megasam}. WildGS-SLAM~\cite{zheng2025wildgs} leverages image-rendering inconsistency from 3D Gaussian Splatting (3DGS)~\cite{kerbl3Dgaussians} to train on-the-fly a per-sequence specialized motion detector, but its effectiveness diminishes when input sequences lack sufficient temporal and spatial coverage.  MegaSaM~\cite{li2024megasam}, on the other hand, trains a more general motion prediction network, offering a promising direction but suffers from a key limitation: it attempts to decode motion from the recurrent hidden state of its flow operator, which is designed to propagate flow information and not to explicitly segment motion. Hence, it lacks the capacity and feature richness required for a complex segmentation task. Furthermore, its reliance on a synthetic training corpus of limited diversity creates a domain gap, causing its performance to degrade on long, real-world trajectories. Critically, existing dynamic-scene methods such as the above often fail to maintain accuracy with solely static inputs. We argue that a robust SLAM system must excel in both static and dynamic scenarios, as real-world applications are inherently unpredictable and will inevitably encounter a mix of both.

We present \textbf{\textit{\project{}}}, a unified and generalizable pose estimation framework capable of handling both dynamic and static environments, while remaining robust across short-term and long-term sequences. Our key insight is to fuse the two powerful paradigms in deep visual pose estimation and 3D reconstruction: the end-to-end optimization of differentiable BA frameworks~\cite{teed2021droid} and the rich perceptual frontend of feed-forward models~\cite{murai2025mast3r}. To this end, we adopt the differentiable BA pipeline of DROID-SLAM~\cite{teed2021droid}, which relies on a learnable update operator to refine poses and geometry. We enhance this pipeline in two fundamental ways.

First, we create a new 3D-aware update operator. We are the first to replace the standard, simple CNN encoder with a powerful, pre-trained 3D-aware feature extractor~\cite{mast3r_eccv24}, providing rich geometric priors. We train this operator on a highly diverse, augmented dataset of both static and dynamic scenes with varied motion patterns, improving robustness to diverse, in-the-wild inputs. Second, we propose a new motion mask detector that also leverages these 3D-aware features. Unlike MegaSaM, which decodes from a weak recurrent state, our detector is a dedicated module receiving multi-level features from the MASt3R backbone. This high-capacity design enables more accurate dynamic region detection and allows seamless integration as a robust weighting in the BA process.

Our full pipeline integrates these components with monocular depth priors for metric scale, along with loop closure and global BA for long-term consistency. Extensive experiments show \project{} consistently outperforms prior work across a wide variety of benchmarks. We further demonstrate that our improved pose accuracy and motion masks are beneficial for downstream tasks like consistent video depth estimation.
In summary, our main contributions are: \begin{itemize} 
\item A unified monocular pose estimator, \textit{\project{}}, that operates robustly in highly dynamic environments while maintaining state-of-the-art performance on static and short-term datasets. 
\item An enhanced update operator for differentiable BA, which leverages 3D-aware features and is trained on a new, diverse curriculum of static and dynamic data. 
\item A novel, high-capacity motion mask detector to identify dynamic regions that offers seamless integration into the BA pipeline. 
\end{itemize}

\section{Related Work}
\label{sec:related_work}
\paragraph{Feed-forward 3D Reconstruction}
Feed-forward 3D reconstruction has emerged as a significant paradigm from classical optimization-based techniques, aiming to solve SfM in a single pass. These methods, pioneered by DUSt3R~\cite{wang2024dust3r} and MASt3R~\cite{leroy2024grounding}, directly predict 3D point maps and camera parameters from a set of unposed images in a single forward pass, followed by VGGT~\cite{wang2025vggt}, which improved scalability and generalization. More recently, the focus has expanded to address the challenge of dynamic environments. MonST3R~\cite{zhang2024monst3r} directly finetunes DUSt3R~\cite{wang2024dust3r} to support robust prediction under dynamic scenes. CUT3R~\cite{cut3r} proposes a recurrent state model that continuously updates its pose and point map prediction with each new observation. $\pi^{3}$~\cite{wang2025pi} and MapAnything~\cite{keetha2025mapanything} directly train multi-view transformers on large-scale static and dynamic datasets to generalize to in-the-wild data. While fast and robust, these models struggle with long sequences, precise camera pose estimation, and are not fully compatible with camera calibration.

\paragraph{Visual SLAM}
Monocular Visual SLAM systems~\cite{zhang2024glorie,sandstrom2024splat,zhu2024nicer,matsuki2024gaussian,teed2021droid} estimate camera motion and scene geometry from RGB sequences. Classical approaches rely on sparse feature correspondences ~\cite{Mur2017TRO, campos2021orb3} or direct photometric consistency~\cite{engel2014lsd, Engel2017DSO, forster2014svo}, which often produce false matches in dynamic regions, leading to tracking failures.   A prominent approach within this domain is the differentiable optimization framework proposed by Droid-SLAM~\cite{teed2021droid}. This paradigm trains a deep neural network to predict intermediate variables, which are consumed by a dense, differentiable BA layer. While this design has demonstrated superior performance in static environments, its network is trained purely on static data, limiting its robustness to dynamic inputs. Furthermore, its frontend consists of a simple CNN-based feature extractor trained from scratch. In contrast, our work leverages the powerful, pre-trained feature backbone from MASt3R~\cite{mast3r_eccv24}, a successful feed-forward SfM model. Another method that leverages MASt3R is Mast3r-SLAM~\cite{murai2025mast3r}, which builds its system bottom-up from the MASt3R network's direct 3D point map outputs. While it achieves robust real-time tracking, it operates as a fully training-free system. This contrasts with our approach, which integrates MASt3R's internal features into a learnable, optimization-based pipeline. Moreover, this approach, like most in its category~\cite{maggio2025vggt_slam}, is primarily designed for static scenes and struggles with significant dynamic elements.

\paragraph{Pose Estimation in Dynamic Scenes}
To address the challenge of pose estimation in dynamic environments, many approaches first detect dynamic objects by leveraging geometric cues~\cite{cheng2019improving,palazzolo2019iros,scona2018staticfusion, goli2025romo, wimbauer2021monorec, zhao2022particlesfm, zhang2022structure}, or by predefining semantic priors to segment out moving objects, as in MASK-SLAM~\cite{kaneko2018mask}, Crowd-SLAM~\cite{soares2021crowd}, and ViPE~\cite{huang2025vipe}. Others combine both strategies~\cite{li2025ddn,schischka2024dynamon, xu2024dg, jiang2024rodyn, bescos2018dynaslam}. However, these semantic-based methods are inherently brittle: they are constrained by predefined categories, fail to detect novel dynamic objects, and are sensitive to segmentation quality. 
Other recent learning-based methods avoid these priors. BA-Track~\cite{Chen_2025_ICCV} learns to decompose the observed point motion into static and dynamic components. Since only the static component is induced by camera motion, it excludes the dynamic motion from bundle adjustment to eliminate distractions from moving objects. WildGS-SLAM~\cite{zheng2025wildgs} integrates dynamic object handling by training an uncertainty MLP online, supervised by image–rendering inconsistencies from 3D Gaussian splatting.
This design, however, makes it reliant on high-quality mapping and wide viewpoint coverage. Furthermore, its pose estimation is tightly coupled with the computationally demanding mapping module and online mask optimization overhead, resulting in a slow pipeline. Alternatively, MegaSaM~\cite{li2024megasam} learns a motion estimator from large-scale training data. However, it suffers from key limitations: its training on a synthetic corpus of limited diversity creates a domain gap, hindering generalization to long sequences. Architecturally, it decodes motion from the ConvGRU's hidden state—a low-capacity representation designed for flow. In contrast, our method trains a separate, high-capacity detector leveraging 3D-aware features~\cite{mast3r_eccv24}. 

\section{\project{}}
\label{sec:method}

\begin{figure*}[ht!]
\centering
 \includegraphics[width=0.9\linewidth]{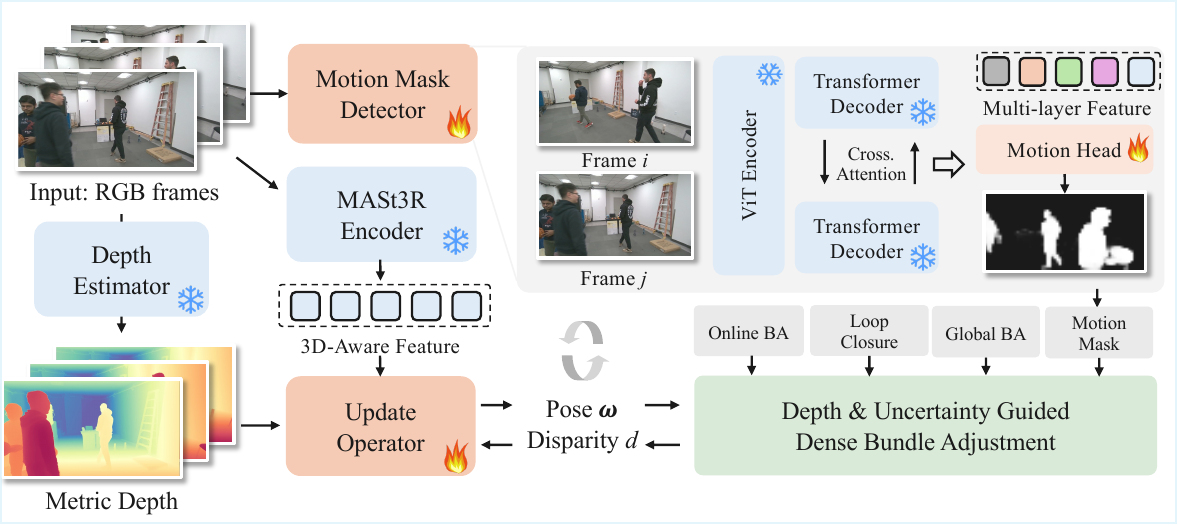}\\
\vspace{-2mm}
\caption{\textbf{System Overview.} \project{} robustly estimates the camera trajectory from a monocular RGB sequence. We leverage 3D-aware features from the frozen MASt3R encoder~\cite{mast3r_eccv24}, which are fed into our update operator 
. Concurrently, a motion mask detector generates motion masks from the backbone's multi-layer features. These outputs, combined with the metric depth prior~\cite{wang2025moge}, enable our Dense Bundle Adjustment layer to mitigate dynamic outliers and perform robust trajectory optimization.}
\vspace{-5pt}
\label{fig:pipeline}
\end{figure*}
Given a monocular RGB sequence $\{I_i\}_{i=1}^{N}$ with known camera intrinsics $(f_x, f_y, c_x, c_y)$, our goal is to estimate the per-frame camera poses $\{\bomega_i\}_{i=1}^{N}\in SE(3)$.
We first review the preliminaries of DROID-SLAM~\cite{teed2021droid}, then describe how we integrate the feed-forward front-end MASt3R~\cite{leroy2024grounding} into a differentiable SLAM framework for end-to-end training. 
We further introduce our learned motion mask detector and present a multi-stage training strategy for all components. 
Finally, we describe how the trained modules are combined for unified inference.

\subsection{Preliminaries}
\label{subsec:droid}
DROID-SLAM~\cite{teed2021droid}
is characterized by two core components: a learnable update operator and a differentiable BA pipeline. The state of each keyframe $i$ comprises the camera pose $\hat{\omega_{i}} \in SE(3)$ and downsampled disparity map $\hat{d_i} \in \mathcal{R}^{\frac{H}{8} \times \frac{W}{8}}$. The optimization process is acted on a frame graph $G$, where edges $(i,j) \in G$ connect keyframes with significant observational overlap.
For each edge in the graph, the update operator iteratively refines intermediate variables which are further utilized in the differentiable BA process to optimize the camera poses and geometry.

\paragraph{Update Operator}
For each image pair $(I_i, I_j)$ forming an edge in the pose graph, we estimate the optical flow by projecting and back-projecting pixels as
$\tilde{\mathbf{f}}_{i,j} = \Pi_c\left( \hat{\bomega}_j^{-1}\hat{\bomega}_i \Pi_c^{-1}\left({p}_i, \hat{d}_i\right)\right)$,
where $\Pi_c$ denotes the pinhole camera projection and $p_i$ is a pixel location in the reference frame $i$. The update operator then iteratively refines the pose and disparity by adjusting intermediate variables guided by the estimated optical flow
\begin{equation}(\hat{\mathbf{f}}^{t+1}_{i,j},\hat{\mathbf{w}}^{t+1}_{i,j},\hat{\mathbf{\eta}^{t+1}},\hat{\mathbf{u}}^{t+1}) = F(I_{i},I{j},\tilde{\mathbf{f}}^{t}_{i,j},\mathbf{r}^{t}_{i,j}),
\end{equation}
where $\hat{\mathbf{f}}^{t}_{i,j}\in \mathcal{R}^{\frac{H}{8} \times \frac{W}{8} \times 2}$ is the predicted optical flow between the two frames $i$ and $j$,  $\hat{\mathbf{w}}^{t}_{i,j}\in \mathcal{R}^{\frac{H}{8} \times \frac{W}{8} \times 2}$ is the confidence of the predicted optical flow, $\hat{\mathbf{\eta}^{t}}$ is a damping factor used to stabilize the BA process, $\hat{\mathbf{u}}^{t} \in \mathcal{R}^{8 \times8}$ is a mask for disparity upsampling, and  $\mathbf{r}^{t}_{i,j} = \hat{\mathbf{f}}^{t}_{i,j} - \tilde{\mathbf{f}}^{t}_{i,j}$. The update operator $F$ comprises a CNN feature encoder, a feature correlation module, and a ConvGRU-based recurrent network to iteratively refine the state variables. 
For more architecture details, please refer to the supplementary material.

\paragraph{Differentiable BA}
The differentiable BA layer jointly optimizes all camera poses $\hat{\mathbf{\omega}}$ and disparity maps $\hat{d}$ by minimizing the reprojection error over the frame graph $G$:
\begin{equation}
E(\hat{\mathbf{\omega}}, \hat{d}) = \sum_{(i, j) \in G} \left\|\hat{\mathbf{f}}_{i j}-\tilde{\mathbf{f}}_{i j}\right\|_{\Sigma_{i j}}^2,
\label{eqa:BA}
\end{equation}
where $\Sigma_{i j}^{-1} = \text{diag}(\hat{\mathbf{w}}_{i,j})$.
This objective penalizes the weighted residual between the predicted flow $\hat{\mathbf{f}}$ and the geometry-induced flow $\tilde{\mathbf{f}}$, weighted by the confidence $\hat{\mathbf{w}}$ from the update operator. This non-linear least-squares problem is solved by the Gauss-Newton algorithm. The cost function is linearized using local parameterization, and the resulting linear system for updates $(\Delta\hat{\mathbf{\omega}}, \Delta\hat{d})$ is solved efficiently by applying the Schur complement. The entire process is fully differentiable, enabling end-to-end training by backpropagating gradients through the solver.


\subsection{Feed-forward Transformer Encoder}
Recent advances in feed-forward models, such as MASt3R~\cite{mast3r_eccv24}, have demonstrated that large-scale pre-training can produce powerful, 3D-aware image representations. In contrast, the update operator in DROID-SLAM~\cite{teed2021droid} employs a relatively simple CNN encoder, trained from scratch, merely to extract auxiliary features for its recurrent ConvGRU. Therefore, we propose to replace this simple CNN with the pre-trained MASt3R backbone, integrating a feed-forward frontend with the robust optimization of a differentiable BA backend.

A direct integration is not possible, as the outputs of the MASt3R backbone are not compatible with the inputs expected by the DROID update operator. The MASt3R architecture consists of a ViT encoder~\cite{dosovitskiy2020image} and two intertwined decoders ($\text{DecBlk}^{1}$, $\text{DecBlk}^{2}$) that produce complex, multi-level correspondence tokens. An obvious design choice is to adapt the final multi-level tokens or the intermediate ViT features. 
However, we find that fusing the decoder tokens with the ConvGRU layer is computationally prohibitive for both training and inference.

Therefore, we opt to use the features from the ViT encoder. To make the encoder compatible with the ConvGRU, we introduce a lightweight adapter module. This module consists of two convolutional residual layers that take the patch-based ViT features as input. It processes these features to produce the local feature maps and global context information that the ConvGRU subsequently uses to update the flow field. We empirically selected MASt3R over other feed-forward models~\cite{wang2025vggt,wang2025pi} as we found its features yielded the best performance. We hypothesize this is because MASt3R's pre-training objectives on point map precision and 2D pixel correspondences are highly aligned with the update operator's role in estimating the optical flow, providing "cleaner" and more relevant features than those from more versatile models.

\subsection{Motion Mask Detector}
\label{subsec:motion_mask}
The BA objective in \eqnref{eqa:BA} only holds on static scenes, where only ego-motion $\tilde{\mathbf{f}}_{ij}$ determines the full optical flow $\hat{\mathbf{f}}_{ij}$. This assumption is violated in dynamic environments. For a pixel $p_i$ corresponding to a moving 3D point, its true optical flow $\tilde{\mathbf{f}}^{\star}_{i,j}$ is a composite of camera motion and the object's self motion:
$$
\tilde{\mathbf{f}}^{\star}_{i,j} = \Pi_c\left( \hat{\mathbf{\omega}}_j^{-1}\hat{\mathbf{\omega}}_i \Pi_c^{-1}\left({p}_i, \hat{d}_i\right) + X_{i,j}(p_i)\right)
$$
where $X_{i,j}(p_i)$ represents the object's 3D displacement between frames $I_i$ and $I_j$. Consequently, directly minimizing \eqnref{eqa:BA} forces the optimizer to incorrectly adjust $\hat{\mathbf{\omega}}$ and $\hat{d}$ to compensate for the large residuals from dynamic pixels, leading to inaccurate state estimation. To mitigate this, we propose a deep motion detector that takes an image pair as input and predicts a motion map to identify and filter these dynamic regions.

Our motion detector, illustrated in \figref{fig:pipeline}, is designed to predict a downsampled motion map $\hat{M}_{i,j} \in \mathbb{R}^{\frac{H}{8} \times \frac{W}{8}}$. This map is used to robustify the BA objective (\eqnref{eqa:BA}) by identifying and down-weighting residuals from dynamic pixels. To achieve this, we adapt the MASt3R backbone~\cite{mast3r_eccv24} by aggregating the multi-level correspondence tokens produced by its transformer decoder via a DPT-style fusion layer~\cite{ranftl2021vision}. The fused features are then passed to a final CNN-based motion head, which regresses the motion map $\hat{M}_{i,j}$ that identifies the dynamic regions on reference frame $i$.

\begin{figure}[tb]
\centering
  \includegraphics[width=1.0\columnwidth]{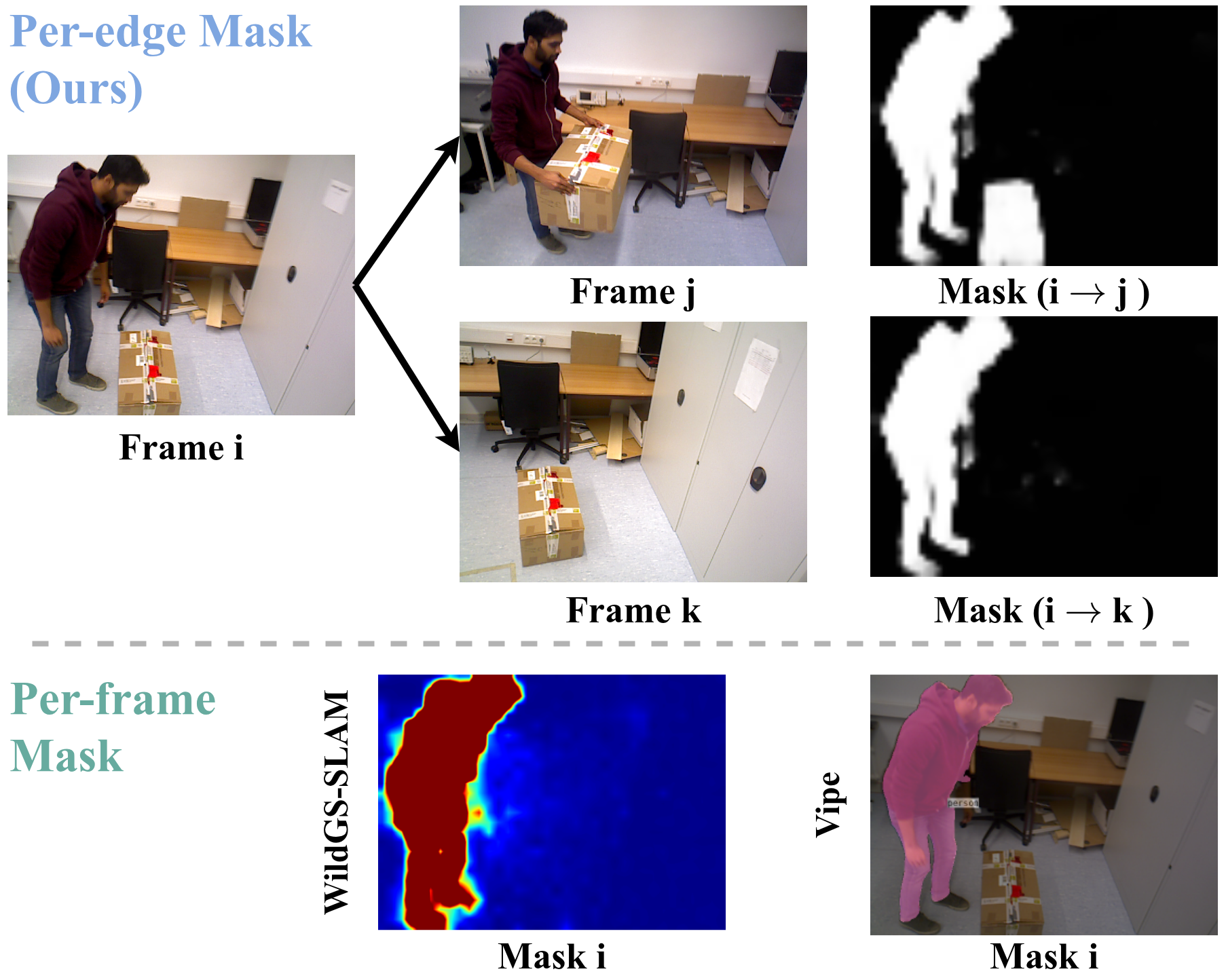} 
    \vspace{-10pt}
    \caption{\textbf{Visualization of Motion Masks.} 
    \textit{Top}: Our per-edge masks (Frame~$i \!\rightarrow\! j$ and $i \!\rightarrow\! k$) resolve temporal ambiguity by capturing motion relative to a second frame, enabling fine-grained detection of inconsistencies along each frame-graph edge.
    \textit{Bottom}: Per-frame masks from prior methods (WildGS-SLAM~\cite{zheng2025wildgs} and Vipe~\cite{huang2025vipe}) are shown for comparison; these approaches produce frame-level predictions that are unable to identify transient motion.}
    \label{fig:masks}
    \vspace{-15pt}
\end{figure}

Our edge-dependent mask design avoids the ambiguity of per-frame~\cite{zheng2025wildgs} or global dynamic object segmentation~\cite{huang2025vipe}. An object may move transiently, remaining static in other parts of the sequence, making a single-frame "dynamic" label unreliable. For instance, as shown in \figref{fig:masks}, a global-consistency-based method~\cite{zheng2025wildgs} fails to identify the moved box, as it is static in most other views. Our pairwise detector circumvents this. It only masks objects if they move between the input pair ($I_i, I_j$), thereby correctly identifying true outliers for a specific edge in the BA graph. If the object is static within that pair, it is preserved as a valid constraint, improving robustness. This design resolves temporal ambiguity and integrates directly into the BA optimization.

\subsection{Training}
Our pipeline contains two learnable modules, the update operator and the motion detector, which we train using a multi-stage curriculum. Unlike prior work~\cite{sandstrom2024splat,zhang2024glorie,zheng2025wildgs} that uses the pre-trained DROID-SLAM operator, our goal is to train a new operator with enhanced robustness to both static and dynamic scenes. This is followed by a separate stage to train the motion detector.

\paragraph{Training Datasets}
We train our methods on a diverse set of synthetic datasets, including the static TartanAir V2~\cite{TartanAirV2} and TartanGround~\cite{patel2025tartanground}, and the dynamic Dynamic Replica~\cite{karaev2023dynamicstereo} and OmniWorld-Game~\cite{zhou2025omniworld}. While these provide various trajectories, they underrepresent certain challenging camera motions. To address this, we augment our training pool with data from the Kubric simulator~\cite{greff2022kubric}. We generate both static and dynamic sequences focusing on three distinct motion patterns: linear translation, pure rotation, and target-locked motion, i.e., fixing the look-at point while the camera moves. The latter two are less prevalent in the other public datasets and enhance the dataset diversity.

\paragraph{Training Strategy}
We first train the update operator end-to-end using only the static datasets. This stage trains the model to effectively learn pairwise flows induced purely by camera ego-motion. The loss function, similar to \cite{teed2021droid}, is a weighted sum:
\begin{equation}
\mathcal{L}_{1} = w_{cam}\mathcal{L}_{cam} + w_{flow}\mathcal{L}_{flow} + w_{res}\mathcal{L}_{res},
\end{equation}
where $\mathcal{L}_{cam}$ is the loss on the camera poses, $\mathcal{L}_{flow}$ compares the geometry-induced flow $\tilde{f}$ with the ground truth flow, and $\mathcal{L}_{res}$ is the residual loss between $\tilde{f}$ and the operator-predicted flow $\hat{f}$. 

In the second stage, we finetune the operator on a mix of static and dynamic datasets. To train the model to handle dynamic content, we provide the annotated motion masks $\mathbf{M}$ and inject them directly into the BA process by modifying the weighting matrix from \eqnref{eqa:BA} to $\bar{\Sigma}_{i j}^{-1} = \text{diag}(\hat{\mathbf{w}}_{i,j}\mathbf{M}_{i})$. This stage trains the update operator to adapt to masked inputs and generalize to dynamic outliers.

Finally, we train the motion mask detector by coupling it with the frozen update operator and the differentiable BA layer. In this stage, we replace the ground truth motion masks with the predicted masks from our motion detector. We remove the flow loss and residual loss from the supervision to focus solely on the pose and mask quality:
\begin{equation}
\mathcal{L}_{2} = w_{cam}\mathcal{L}_{cam} + w_{mask}\mathcal{L}_{BCE},
\label{eqn:mask_loss}
\end{equation}
where $\mathcal{L}_{BCE}$ is the binary cross-entropy loss between the predicted motion map and the annotated mask.
\begin{table*}[t]
\centering
\footnotesize
\setlength{\tabcolsep}{4.5pt}
{
\begin{tabular}{lrrrrrrrrrrrr}
\toprule
Method & \texttt{ANYmal1} & \texttt{ANYmal2} & \texttt{Ball} & \texttt{Crowd} & \texttt{Person} & \texttt{Racket} & \texttt{Stones} & \texttt{Table1} & \texttt{Table2} & \texttt{Umbrella} & Avg. \\
\midrule

\multicolumn{12}{l}{\cellcolor[HTML]{EEEEEE}{\textit{Keyframe Poses}}} \\
MASt3R-SLAM~\cite{murai2025mast3r} & \rd 0.5 & \rd1.2 & \rd0.5 & 1.3 & 4.5 & 1.9 & \rd 0.9 & 4.2 & 29.5 & 1.5 & 4.59\\
VGGT~\cite{wang2025vggt} & 1.2 & 87.2 & 1.0 & 1.3 & 2.9 & 1.7 & 2.0 & 1.9 & 58.8 & 1.2 & 15.91 \\
$\pi^{3}$~\cite{wang2025pi} & 0.8 & 3.7 & 0.9 & 1.1 & 1.8 & \rd 1.5 & 1.3 & 2.4 & \rd 5.7 & 1.1 & \rd 2.02\\

\hdashline
\noalign{\vskip 1pt}
\multicolumn{12}{l}{\cellcolor[HTML]{EEEEEE}{\textit{Full Trajectory}}} \\
DROID-SLAM~\cite{teed2021droid} & 0.6 & 4.7 & 1.2 & 2.3 & \nd 0.6 & 1.5 & 3.4 & 48.0 & 95.6 & 3.8 & 16.17\\
WildGS-SLAM~\cite{zheng2025wildgs} & \fs 0.2 & \fs 0.3 & \fs 0.2 & \fs 0.3 & \rd 0.8 & \fs 0.4 & \fs 0.3 & \nd 0.6 & \nd 1.3 & \fs 0.2 & \nd 0.46\\
MegaSaM~\cite{li2024megasam} & 0.6 & 2.7 & 0.6 & \rd 1.0 & 3.2 & 1.6 & 3.2 & \rd 1.0 & 9.4 & \rd 0.6 & 2.40\\
ViPE~\cite{huang2025vipe} & \nd 0.4 & \nd 0.7 & \nd 0.4 & \nd 0.6 & 4.8 & \rd 1.2 & \nd 0.7 & 1.7 & 14.8 & \nd 0.5 & 2.59\\
{\textbf{\project{} (\ours)}} & \fs 0.2 & \fs 0.3 & \fs 0.2 & \nd 0.4 & \fs 0.4 & \nd 0.5 & \fs 0.3 & \fs 0.5 & \fs 1.0 & \fs 0.2 & \fs 0.39\\

\bottomrule
\end{tabular}

}
\vspace{-5pt}

\caption{\textbf{Tracking Performance on Wild-SLAM MoCap Dataset~\cite{zheng2025wildgs}} (ATE RMSE $\downarrow$ [cm]). Best results are highlighted as\colorbox{colorFst}{\bf first},\colorbox{colorSnd}{second}, and\colorbox{colorTrd}{third}. All baseline methods were run using their publicly available code.} 
\label{tab:mocap_tracking}
\vspace{-10pt}
\end{table*}
\subsection{Inference}
During inference, our system operates as an online visual odometry. We designate a new incoming frame as a keyframe if it has sufficient motion \wrt the latest keyframe. For each new keyframe $I_i$, we initialize its disparity $\hat{d}_i$ using the metric depth $D_i$ from Moge2~\cite{wang2025moge} and form new edges with its co-visible keyframes in the graph $G$. For each new edge $(i, j)$ created in the graph, we run our motion detector to predict the motion mask $M_{i,j}$. The poses and disparities are then optimized via BA by minimizing a composite objective similar to the training stage:
\begin{equation}
E(\hat{\mathbf{\omega}}, \hat{d}) = \sum_{(i, j) \in G} \left\|\hat{\mathbf{f}}_{i j}-\tilde{\mathbf{f}}_{i j}\right\|_{\bar{\Sigma}_{i j}}^2 + \lambda \sum_{i \in G}\left\|\left(\hat{d}_i-1 / D_i\right)\right\|^2
\label{eqa:BA_infer}
\end{equation}
where the reprojection error is further weighted by the predicted motion mask, $\bar{\Sigma}{i j}^{-1} = \text{diag}(\hat{\mathbf{w}}{i,j}\hat{M}{i,j})$, with an additional disparity regularization term which encourages the estimated disparity to cohere with the metric depth prior.
This optimization is performed in a sliding window manner, where old keyframes and edges are removed when new keyframes come in.

To mitigate scale and pose drift in longer sequences, we employ loop closure and global BA in addition to the local window optimization, similar to \cite{sandstrom2024splat,zheng2025wildgs,zhang2024glorie}. We detect loops by computing optical flow between current active keyframes and past keyframes; if the temporal interval between two frames exceeds a threshold $\tau$ and they have sufficient co-visibility, we create a new graph with these loop closure edges and perform BA on them. Global BA is performed on all keyframes to ensure global consistency, running periodically during the sequence with fewer iterations and as a final post-processing step. In the final global BA, we remove the second term in \eqnref{eqa:BA_infer}, which is the depth regularizer term. We empirically find that while the metric depth provides good initialization, its inherent noise can be detrimental when estimates are highly refined. 

\section{Experiments}
\label{sec:exp}

\subsection{Experimental Setup}
\paragraph{Datasets}
We evaluate our approach on a wide range of datasets to assess its performance under various conditions.
\begin{itemize}
\item \textbf{Dynamic Scenes}: To demonstrate robustness to dynamic objects, we follow the benchmark from \cite{zheng2025wildgs} and evaluate on Wild-SLAM MoCap Dataset~\cite{zheng2025wildgs}, Bonn RGB-D Dynamic Dataset~\cite{palazzolo2019iros}, and the dynamic sequences in TUM RGB-D Dataset~\cite{sturm2012benchmark}.

\item \textbf{Low-Motion and Short Trajectories}: To assess performance on challenging sequences with minimal egomotion, we evaluate on the MPI Sintel dataset~\cite{butler2012naturalistic}. This dataset contains 18 short sequences, each of which has 20-50 frames, with small camera motion.

\item \textbf{Static Scenes}: Finally, to validate performance on standard static benchmarks, we evaluate on 7-Scenes Dataset~\cite{shotton2013scene} and the nine static sequences in TUM RGB-D Dataset~\cite{sturm2012benchmark}, following \cite{murai2025mast3r}. We also report the evaluation on the ScanNet Dataset~\cite{dai2017scannet} in the supplementary material.
\end{itemize}

\paragraph{Metrics}
We follow the standard monocular SLAM evaluation pipeline~\cite{zhang2024glorie,zheng2025wildgs,murai2025mast3r}.
The estimated trajectory is aligned to the ground truth using Sim(3) Umeyama alignment~\cite{umeyama1991least}, after which we compute the RMSE of the absolute trajectory error (ATE).

\paragraph{Implementation details}
The update operator is trained in a two-stage process using the Adam optimizer. The first pre-training stage uses only static datasets and runs for 200K iterations with a learning rate (LR) of $1.5\times10^{-3}$. We use a batch size of 3, where each sample consists of a 7-frame video sequence. The second stage is conducted jointly on both static and dynamic datasets for 100K iterations with a reduced LR of $1\times10^{-4}$. Subsequently, the motion mask detector is trained for 15K iterations with a batch size of 8 and a learning rate of $5\times10^{-3}$. The entire training curriculum takes approximately two weeks on 8 A100 GPUs. We employ gradient norm clipping with a threshold of 2.5 to ensure training stability. All training images are resized to $384\times512$. During inference, we resize input images to a similar resolution while preserving the original aspect ratio.

\begin{table*}[t]
\centering
\footnotesize
\setlength{\tabcolsep}{4.5pt}
\begin{tabular}{lrrrrrrrrr}
\toprule
Method & \texttt{Balloon} & \texttt{Balloon2} & \texttt{Crowd} & \texttt{Crowd2} & \texttt{Person} & \texttt{Person2} & \texttt{Moving} & \texttt{Moving2} & Avg.\\
\midrule
\multicolumn{10}{l}{\cellcolor[HTML]{EEEEEE}{\textit{Keyframe Poses}}} \\
MASt3R-SLAM~\cite{murai2025mast3r} & \rd 2.8 &3.1 & 4.4 & 6.6 & \fs 2.6 & \fs 1.9 & 7.5 & 2.8 & 3.96\\
VGGT~\cite{wang2025vggt} & 3.3 & 3.5 & 2.0 & \nd 2.1 & 6.6 & 27.1 & \nd 1.4 & 4.0 & 6.26\\
$\pi^{3}$~\cite{wang2025pi} & \fs 1.5 & \fs 1.7 & 2.7 & 2.9 & 3.6 & 7.3 & \nd 1.4 & \nd 2.3 & \rd2.92\\

\hdashline
\noalign{\vskip 1pt}
\multicolumn{10}{l}{\cellcolor[HTML]{EEEEEE}{\textit{Full Trajectory}}} \\
DROID-SLAM~\cite{teed2021droid} & 7.5 & 4.1 & 5.2 & 6.5 & 4.3 & 5.4 &  2.3 & 4.0 & 4.91\\
WildGS-SLAM~\cite{zheng2025wildgs} & \rd 2.8 & \nd 2.4 & \nd 1.5 & \rd 2.3 & \nd 3.1 & \rd 2.7 & \rd 1.6 & \fs 2.2 & \fs 2.31\\
MegaSaM~\cite{li2024megasam} & 3.7 & \rd 2.6 & \rd 1.6 & 7.2 & 4.1 & 4.0 & \nd 1.4 & 3.4 &
3.51\\
ViPE~\cite{huang2025vipe} & 3.3 & 2.7 & 2.6 & 3.9 & \rd 3.4 & 3.8 & \nd 1.4 & \nd 2.3 & 2.93\\
{\textbf{\project{} (\ours)}} & \nd 2.6 &  \nd2.4 & \fs 1.4 & \fs 1.9 & 4.0 & \nd 2.6 & \fs 1.3 & \rd 2.7 & \nd 2.36\\

\bottomrule
\end{tabular}

\vspace{-2mm}
\caption{\textbf{Tracking Performance on Bonn RGB-D Dynamic Dataset~\cite{palazzolo2019iros}} (ATE RMSE $\downarrow$ [cm]). 
} 
\label{tab:bonn_tracking}
\vspace{-10pt}
\end{table*}

\begin{table}[t]
\centering
\footnotesize
\resizebox{\columnwidth}{!}
{
\begin{tabular}{lrrrrrrrr}
\toprule
Method & \texttt{f3/ws} & \texttt{f3/wx} & \texttt{f3/wr} & \texttt{f3/whs} & Avg.\\
\midrule
\multicolumn{6}{l}{\cellcolor[HTML]{EEEEEE}{\textit{Keyframe Poses}}} \\
MASt3R-SLAM~\cite{murai2025mast3r} & 0.7 & 1.7 & 3.9 & 3.7 & 2.51\\
VGGT~\cite{wang2025vggt} & 0.7 & \nd 1.5 & 3.6 & 2.0 & 1.96\\
$\pi^{3}$~\cite{wang2025pi} & 0.9 & 2.4 & 3.8 & 2.6 & 2.44\\

\hdashline
\noalign{\vskip 1pt}
\multicolumn{6}{l}{\cellcolor[HTML]{EEEEEE}{\textit{Full Trajectory}}} \\
DROID-SLAM~\cite{teed2021droid} & 1.2 & \rd 1.6 & 4.0 & 2.2 & 2.25\\
WildGS-SLAM~\cite{zheng2025wildgs} & \fs 0.4 & \fs 1.3 & 3.3 & \fs 1.6 & \rd 1.63\\
MegaSaM~\cite{li2024megasam} & \rd 0.6 & \nd 1.5 & \fs 2.6 & \rd 1.8 & \rd 1.63\\
ViPE~\cite{huang2025vipe} & \nd 0.5 & \fs 1.3 & \rd 3.0 & \fs 1.6 & \nd 1.58 \\
{\textbf{\project{} (\ours)}} & \rd 0.6 & \fs 1.3 & \nd 2.7 & \nd 1.7 & \fs 1.57 \\

\bottomrule
\end{tabular}

}
\vspace{-2mm}
\caption{\textbf{Tracking Performance on TUM RGB-D (dynamic) Dataset~\cite{sturm2012benchmark}} (ATE RMSE $\downarrow$ [cm]). 
\vspace{-2mm}
} 
\label{tab:tum_tracking}
\end{table}

\paragraph{Baselines}
We compare \project{} against SOTA methods, categorized as dynamic, static, and feed-forward: WildGS-SLAM~\cite{zheng2025wildgs}, ViPE~\cite{huang2025vipe}, MegaSaM~\cite{li2024megasam}, DROID-SLAM~\cite{teed2021droid}, MASt3R-SLAM~\cite{murai2025mast3r}, VGGT~\cite{wang2025vggt}, and $\pi^{3}$~\cite{wang2025pi}. Several baselines require specific evaluation protocols. For ViPE, we pre-identify and provide moving semantics as input, in accordance with its requirements. MASt3R-SLAM is evaluated only on keyframes, as it does not predict a full trajectory. Similarly, due to high GPU memory consumption on long sequences, VGGT and $\pi^{3}$ are also evaluated only on the keyframe poses identified by our method, except on Sintel, where the full trajectory is used due to the small number of frames. Noting that the keyframe-only evaluation is inherently less challenging than the full-trajectory evaluation employed for our method and other baselines.

\subsection{Pose Estimation}
\paragraph{Wild-SLAM Mocap}
As shown in \tabref{tab:mocap_tracking}, our method achieves the best performance across all the sequences of Wild-SLAM Mocap Dataset~\cite{zheng2025wildgs}. While ViPE~\cite{huang2025vipe} performs comparably on most sequences, its error increases significantly on the \texttt{Table1} and \texttt{Table2} sequences. These scenes contain a human pushing a table while navigating, a challenging scenario for ViPE as tables are typically treated as static and thus not identified as dynamic by its semantic segmentation, leading to tracking errors. MegaSaM~\cite{li2024megasam} and WildGS-SLAM~\cite{zheng2025wildgs} also detect motions without semantic priors. Our method still surpasses them, which we attribute to our use of 3D-aware features~\cite{mast3r_eccv24} and carefully designed training curriculum. Although $\pi^{3}$ is evaluated only on keyframes, its performance is nonetheless lower than ours. This suggests that while feed-forward models excel at dense pointmap predictions, their end-to-end pose estimation accuracy still lags behind optimization-based methods. Lacking mechanisms to handle dynamic content, other static methods perform dramatically worse on the dynamic dataset.

\paragraph{Bonn and TUM Dynamic}
Tracking results on the Bonn RGB-D~\cite{palazzolo2019iros} and TUM Dynamic dataset~\cite{sturm2012benchmark}, presented in \tabref{tab:bonn_tracking} and \tabref{tab:tum_tracking}, demonstrate superior or competitive performance of our method (second-best on Bonn, best on TUM). The Bonn dataset features non-trivial illumination changes across frames. WildGS-SLAM outperforms our approach on this dataset, likely because its explicit handling of illumination changes. \project{} outperforms all other baselines by a clear margin.

\begin{table}[t]
\begin{center}
\small
\resizebox{\columnwidth}{!}
{
\begin{tabular}{lrrr}
\toprule
\multirow{2}{*}{Method}  & Sintel~\cite{butler2012naturalistic} & TUM RGB-D~\cite{sturm2012benchmark} & 7-scenes~\cite{shotton2013scene} \\
 & low-motion & static & static\\
\midrule
\multicolumn{4}{l}{\cellcolor[HTML]{EEEEEE}{\textit{Keyframe Poses}}} \\
MASt3R-SLAM~\cite{murai2025mast3r} &- & \nd0.030 & \fs 0.047\\
VGGT~\cite{wang2025vggt} & 0.072\rlap{$^{\dagger}$}  & 0.042 & 0.058 \\
$\pi^{3}$~\cite{wang2025pi} & 0.047\rlap{$^{\dagger}$}  & 0.038 & 0.052\\

\hdashline
\noalign{\vskip 1pt}
\multicolumn{4}{l}{\cellcolor[HTML]{EEEEEE}{\textit{Full Trajectory}}} \\
DROID-SLAM~\cite{teed2021droid} & 0.030  & 0.038 & \nd0.049\\
WildGS-SLAM~\cite{zheng2025wildgs} & 0.049 & \rd0.037 & 0.051\\
MegaSaM~\cite{li2024megasam} & \nd 0.018 & 0.066 & 0.056\\
ViPE~\cite{huang2025vipe} &  \rd 0.028 & 0.065 & \rd0.050 \\
{\textbf{\project{} (\ours)}} & \fs 0.017\rlap{$^{\star}$} & \fs0.027 & \nd0.049 \\

\bottomrule
\end{tabular}
}
\caption{\textbf{Tracking Performance on Low-motion and Static benchmarks.} Reported as ATE RMSE $\downarrow$ in meters, except for Sintel where trajectories are scale-normalized following MegaSaM~\cite{li2024megasam}.
${\dagger}$ indicates evaluation on all the frames. 
}
\label{tab:pose_others} 
\end{center} 
\vspace{-24 pt}
\end{table}
\paragraph{Sintel}
We report tracking performance on Sintel~\cite{butler2012naturalistic} in \tabref{tab:pose_others}. As these sequences are short (<50 frames), there are no memory constraints on VGGT~\cite{wang2025vggt} and $\pi^{3}$~\cite{wang2025pi}, allowing evaluation on all frames. 
Notably, WildGS-SLAM, which performs competitively on the long dynamic sequences, struggles here, exhibiting higher error than even some static methods. This is because the short sequences are insufficient to build a high-quality 3DGS map for training its uncertainty MLP effectively.

\begin{figure*}[tb]
\centering
\begin{tabular}{cccccc}
   
    \multicolumn{6}{c}{%
        \includegraphics[width=\linewidth]{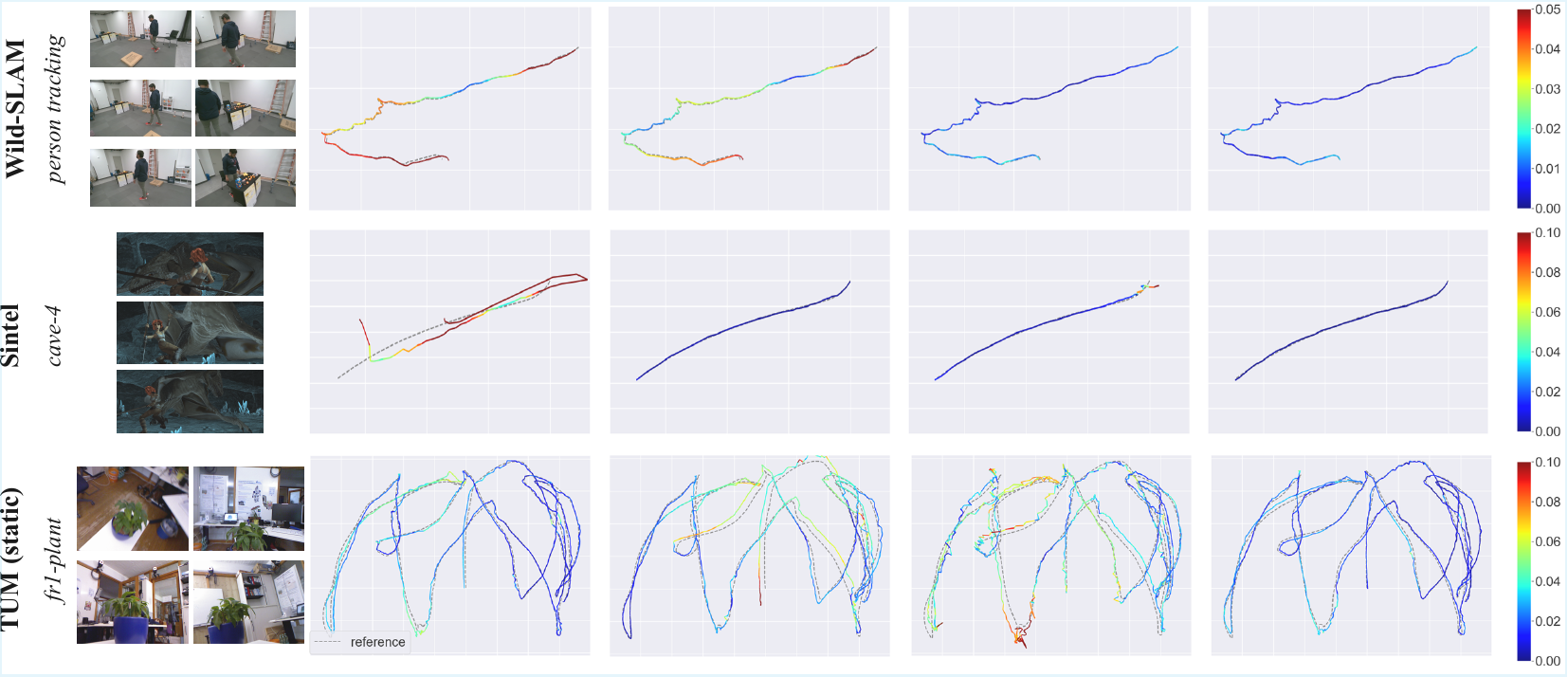}%
    } \\
     \hspace{1.8cm} \small Input & 
    \hspace{1.7cm}\small  ViPE~\cite{huang2025vipe}& 
    \hspace{1.1cm}\small  MegaSAM~\cite{li2024megasam} &
    \hspace{.6cm}\small  WildGS-SLAM~\cite{zheng2025wildgs}& 
    \hspace{0.6cm}\small  \project{} (Ours)\\ 
\end{tabular}  
    \caption{\textbf{Visualization of Camera Trajectories.} The estimated trajectory is colorized by the translation error (ATE).}
    \label{fig:traj}
\end{figure*}

\paragraph{Static Benchmarks}
We report the performance on static datasets in \tabref{tab:pose_others}. Our method achieves the best performance among all full-trajectory pose estimators on both TUM~\cite{sturm2012benchmark} and 7-Scenes~\cite{shotton2013scene}. This highlights the key advantage of \project: while MegaSaM~\cite{li2024megasam} and ViPE~\cite{huang2025vipe} are robust on some dynamic benchmarks, their performance degrades significantly on static datasets, unlike our method which excels at both.

\paragraph{Qualitative Results}
We provide qualitative trajectory comparisons in \figref{fig:traj}, visualizing 2D projections from three representative sequences, one each for dynamic, low-ego-motion, and static. In all examples, our estimated trajectory demonstrates the closest alignment to the ground truth.



\begin{table}[t]
\begin{center}
\small
\resizebox{\columnwidth}{!}
{
\begin{tabular}{lrrr }
\toprule
Method  & Abs.Rel. $\downarrow$ & Log-RMSE $\downarrow$ & $\delta_{1.25}$ $\uparrow$
\\ 
\midrule
DA-v2~\cite{yang2024depth}  & 0.16 & 0.24 &91.1\\
PPD~\cite{xu2025pixel} &0.15 & 0.24 & 94.6 \\
DepthCrafter~\cite{hu2024depthcrafter}  & 0.19 & 0.26 &86.8  \\
Video-Depth-Anything~\cite{chen2025video}& 0.14 & 0.23 & 95.7\\
\midrule
MegaSaM~\cite{li2024megasam}  & 0.13 & 0.23 &94.5\\
{\textbf{\project{} (\ours)}} & \textbf{0.12} & \textbf{0.22} &\textbf{96.3}\\
\bottomrule
\end{tabular}
}
\vspace{-0.5em}
\caption{\textbf{Comparison on Long-video Depth Estimation on Bonn RGB-D Dynamic~\cite{palazzolo2019iros} Dataset.} 
} \label{table:sintel_depth} 
\vspace{-2.0em}
\end{center} 
\end{table}

\subsection{Downstream Application: Depth Estimation}
We demonstrate the benefit of our high-quality poses on video depth estimation. We adopt the depth optimization pipeline from MegaSaM~\cite{li2024megasam}, but substitute their poses with our own and incorporate our predicted motion masks as additional optimization weights. We evaluate on all eight sequences of the Bonn Dynamic Dataset~\cite{palazzolo2019iros}, each of which is temporally subsampled by a stride of 10. The results in \tabref{table:sintel_depth} further showcase our state-of-the-art performance on depth estimation in long sequences, outperforming all baselines on all metrics. This confirms that depth estimation benefits from more accurate pose estimation by \project{}.

\subsection{Ablation Study}
We validate our primary design choices in \tabref{tab:ablation}, where we ablate three key components: (1) the second-stage finetuning of the update operator with mixed static and dynamic datasets (\textit{Mix. Ft.}), (2) the application of our motion mask detector to down-weight dynamic regions during BA (\textit{Mot. Mask}), and (3) the removal of the depth regularizer in the final global BA (\textit{GBA Dep. off}). Our full model, integrating all components, achieves the best performance, validating our complete system design. While dynamic finetuning of the update operator provides significant initial gains, integrating the motion mask detector consistently boosts performance, demonstrating the necessity of explicit motion masks.

\begin{table}[t]
    \footnotesize
    \centering	
    \resizebox{\columnwidth}{!}{
    \begin{tabular}{cccccc}
    \toprule
    \textit{Mix. Ft.} & \textit{Mot. Mask}& \textit{GBA Dep. Off} &  {Wild-SLAM~\cite{zheng2025wildgs}} & {Bonn~\cite{palazzolo2019iros}} & {TUM~\cite{sturm2012benchmark}}  \\
    \midrule
    \noo & \noo & \yess & 2.16 & 5.58 &  2.10\\
    \yess & \noo & \yess & 0.76 & 2.61 &  1.58\\
    \noo & \yess & \yess & 0.61 & 2.55 &  1.66\\
    \yess & \yess & \noo & 2.34 & 2.60 &  2.13\\
    \yess & \yess & \yess & \textbf{0.39} &  \textbf{2.50} & \textbf{1.54} \\
  
    \bottomrule
    \end{tabular}
    } 
    \caption{\textbf{\project{} Ablation Study} (ATE RMSE $\downarrow$ [cm]). We report the average tracking error for each dataset. \textit{Mix. Ft.} denotes finetuning the update operator with the mix of static and dynamic datasets, \textit{Mot. Mask} denotes the motion mask detector, and \textit{GBA Dep. Off} denotes removing the depth regularization term during the final global BA.} 
    \label{tab:ablation}
\vspace{-4mm}
\end{table}
\section{Conclusion}
\label{sec:conclusions}

We presented \project{}, a novel pose estimation framework that achieves robust performance on both static and dynamic scenes. Our method enhances the differentiable BA pipeline by integrating a 3D-aware feature backbone from MASt3R, which powers both a new update operator and a high-capacity motion mask detector. This hybrid design proves superior to prior work, setting a new state-of-the-art on a wide array of static and dynamic datasets and benefiting downstream tasks like depth estimation. We discuss our limitations and future work in the supplementary material.

\paragraph{Acknowledgment}
This work is supported by the Center for Integrated Facility Engineering (CIFE) and the Stanford Robotics Center (SRC). We also thank Stanford’s Marlowe computing clusters and Stanford Sherlock clusters for providing GPU computing for model training and evaluation.


{
    \small
    \bibliographystyle{ieeenat_fullname}
    \bibliography{main}
}
\clearpage
\setcounter{page}{1}
\maketitlesupplementary

\begin{abstract}
In the supplementary material, we provide additional details about the following:
\begin{enumerate}
    \item More information about the training dataset (\secref{sec:dataset}).
    \item Implementation details of \project{}, including more training details and model architecture (\secref{sec:implementation}).
    \item Additional results and discussion (\secref{sec:add_results}).
    \item Limitations and future work (\secref{sec:limitation}).
\end{enumerate}
\end{abstract}

\section{Training Dataset}
\label{sec:dataset}
We trained our model on four publicly available datasets supplemented with data that we generated using the Kubric simulator~\cite{greff2022kubric}. The training datasets encompass both static and dynamic environments. A comprehensive list of the datasets we used is provided in \tabref{tab:dataset}.

While the TartanAir V2~\cite{TartanAirV2} and TartanGround~\cite{patel2025tartanground} datasets primarily feature static scenes, we manually identify dynamic content in 8 of their 65 environments. Among these, ground truth motion masks can be reliably extracted via semantic segmentation for only four specific environments: \texttt{AmusementPark}, \texttt{SeasonalForestAutumn}, \texttt{WaterMillDay}, and \texttt{WaterMillNight}. The dynamic content in the remaining four scenes (\texttt{CarWelding}, \texttt{CyberPunk Downtown}, \texttt{Ocean}, and \texttt{SoulCity}) is complex and difficult to segment accurately. Hence, these four scenes are discarded from the training set. Furthermore, sequences containing significant photographic or geometric artifacts (e.g., camera trajectories passing directly through walls in sequence P004 of \texttt{OldBrickHouseDay}) are also discarded from the training set to maintain data quality.

The OmniWorld-Game dataset~\cite{zhou2025omniworld} provides foreground masks generated using Grounding DINO~\cite{liu2024grounding} to identify bounding boxes, which are then fed as prompts to SAM~\cite{kirillov2023segment} to extract fine masks. However, these generated masks are often noisy, exhibiting incomplete segmentation of instances or entirely missing dynamic objects. Therefore, we do not use their annotated masks as ground truth for calculating the BCE loss in \eqnref{eqn:mask_loss} when training the motion mask detector, but we incorporate them in the second stage of update operator finetuning to encourage robustness against noisy motion maps.

We utilized the Kubric simulator~\cite{greff2022kubric} to generate diverse motion patterns that are under-represented in the public datasets. We considered three specific camera trajectory types: (1) Linear Translation, defined by a linear movement between two randomly selected waypoints; (2) Pure Rotation, involving zero translation, with trajectories encompassing rotation along single or multiple axes; and (3) Target-Locked Motion, where the camera's translation varies but its rotation is constrained to maintain focus on a fixed point. The third type included four distinct translation patterns: lateral, vertical, orbital, and spiral movement. For both static and dynamic scene configurations, we generated a total of 5,500 sequences, each comprising 24 or 36 frames.

\begin{table}[t]
    \footnotesize
    \centering	
    \resizebox{\columnwidth}{!}{
    \begin{tabular}{cccc}
    \toprule
    Dataset Name & Dynamic? & Dynamic Mask? &  Camera Motion Trajectory  \\
    \midrule
    TartanAir V2~\cite{TartanAirV2} & $\text{No}^{\star}$ & - & Drone-style \\
    TartanGround~\cite{patel2025tartanground} & $\text{No}^{\star}$ & - & Grounded Robot \\
    Dynamic Replica~\cite{karaev2023dynamicstereo} & Yes & Yes & Handheld 6-DOF \\
    OmniWorld-Game~\cite{zhou2025omniworld} & Yes & Noisy & Player-controlled \\
    Kubric (Generated)~\cite{greff2022kubric} & Mixed & Yes & Specified motion patterns \\
    \bottomrule
    \end{tabular}
    } 
    \caption{\textbf{Training Datasets.} We utilized four public data sources and generated data from Kubric~\cite{greff2022kubric} for training \project{}. The table details the scene type, the availability and quality of ground truth dynamic masks, and the type of camera motions. OmniWorld-Game~\cite{zhou2025omniworld} uses open-source segmentation models~\cite{kirillov2023segment,liu2024grounding} to generate dynamic mask, hence noisy. While most environments in TartanAir V2~\cite{TartanAirV2} and TartanGround~\cite{patel2025tartanground} are static, there are eight environments containing dynamic objects.}

\label{tab:dataset}
\vspace{-3mm}
\end{table}

\section{Implementation Details}
\label{sec:implementation}

\begin{figure*}[t]
\centering
 \includegraphics[width=0.99\linewidth]{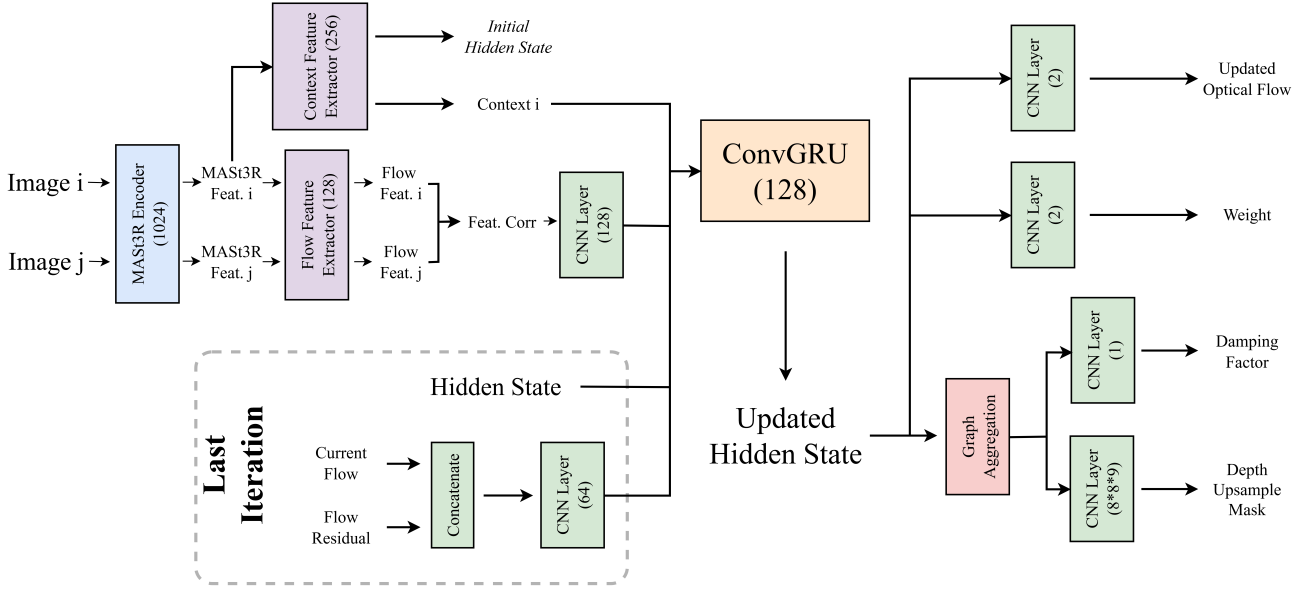}\\
\caption{\textbf{Architecture of Update Operator.} The ConvGRU iteratively updates the hiddens state from the image feature correlation, context features, and the current optical flow. The updated hidden state is further decoded to variables that will be used to guide pose and disparity estimation in the differentiable BA process.}
\label{fig:update}
\end{figure*}

\begin{figure}[t]
\centering
 \includegraphics[width=0.99\linewidth]{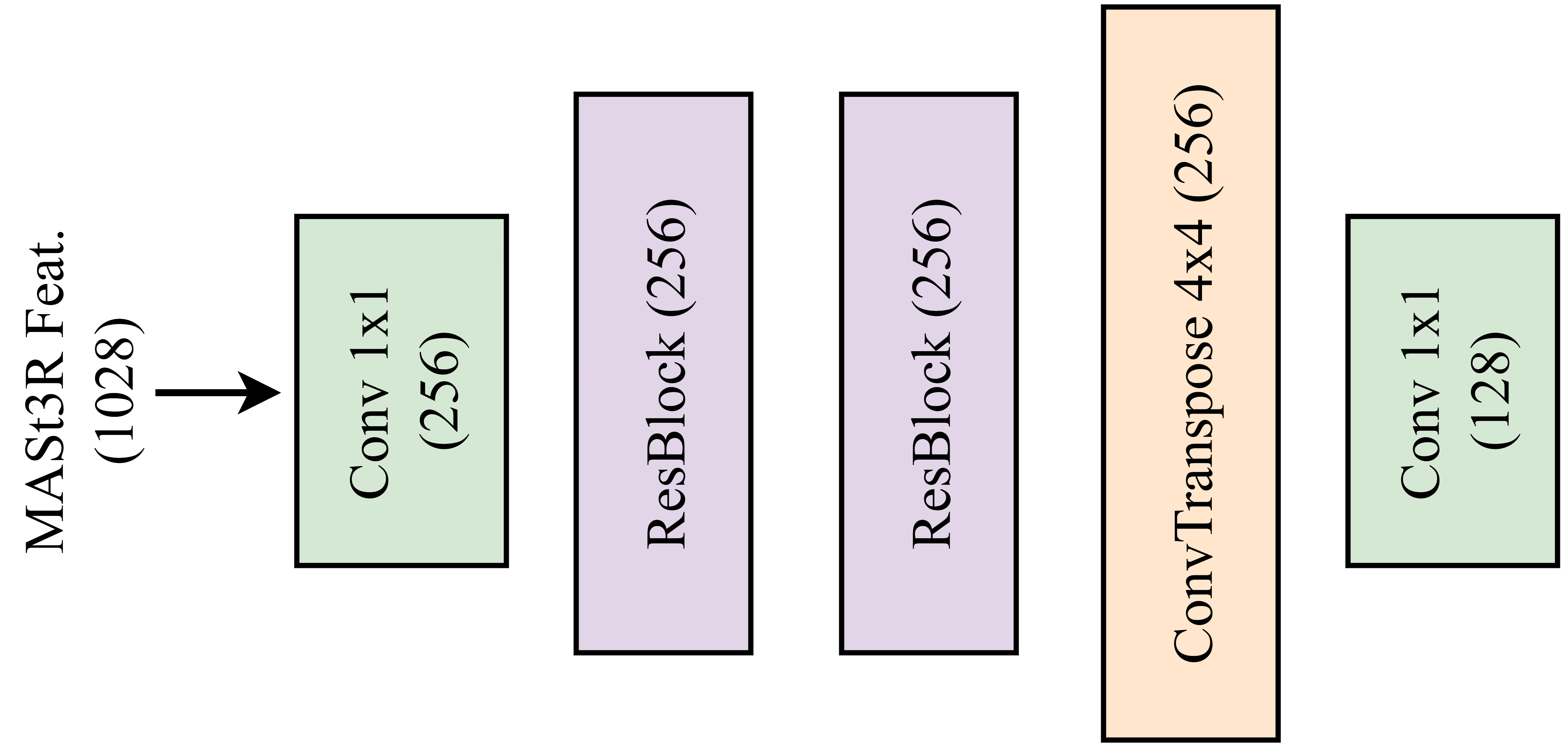}\\
\caption{\textbf{Architecture of the flow feature and context encoders.} Both encoders take the MASt3R features as input and output features at 1/8 of the image resolution. For the context encoder, the dimension of the last convolution layer is 256.}
\label{fig:feat_extractor}
\end{figure}

\subsection{Additional Training Details}
Following~\cite{teed2021droid}, we sample 7 frames per batch from the training sequence. We constrain the average optical flow magnitude between neighboring pairs to fall within the range of 8 to 96 pixels. For all frames, we apply standard data augmentation, comprising photometric transformations (color jitter, grayscale, and Gaussian blurring) and spatial randomization via center resizing. The images are then cropped to the fixed input resolution of $384 \times 512$. Subsequently, a frame graph is constructed over the 7 frames with edges determined by flow distance. We execute 15 iterations of the update operator and differentiable BA.

The total training objective for the update operator combines three weighted loss components. The Pose Loss $\mathcal{L}_{cam}$ is the geodesic distance computed by mapping the relative pose error $\mathbf{G}_{\text{rel}} \mathbf{P}_{\text{rel}}^{-1}$ to the Lie Algebra $\mathfrak{se}(3)$ using the $\text{log}$ map. Here, $\mathbf{G}_{\text{rel}}$ and $\mathbf{P}_{\text{rel}}$ denote the ground truth and predicted relative poses, respectively. $\mathcal{L}_{\text{cam}}$ is defined as the average of the $\text{L}_2$ norms of the translational $\tau$ and rotational $\phi$ components. We further utilize a Flow Loss $\mathcal{L}_{\text{flow}}$ defined by the $L_2$ norm of the difference between the ground truth optical flow and the flow induced by the estimated pose and disparity $\tilde{\mathbf{f}}$. Lastly, we use the Residual Loss $\mathcal{L}_{\text{res}}$, which is calculated by the $L_1$ distance between $\tilde{\mathbf{f}}$ and the flow predicted by the update operator $\hat{\mathbf{f}}$. All three losses are applied across the 15 BA steps, utilizing an increasing temporal weighting scheme $w_k = \gamma^{(15-k)}$, where $\gamma=0.9$ and $k$ is the iteration step.

When training the motion mask detector, the parameters of the update operator are frozen. As mentioned in the main paper, we integrate the predicted motion mask as an additional weight during the 15 BA steps. We retain the Pose Loss $\mathcal{L}_{cam}$, exclude $\mathcal{L}_{\text{flow}}$ and $\mathcal{L}_{\text{res}}$, and add the BCE loss $\mathcal{L}_{\text{BCE}}$, which is only applied when reliable annotated ground truth motion masks are available.

For the initial static pretraining stage of the update operator, we utilize only the static portions of the TartanAir V2, TartanGround, and Kubric-Generated datasets. All remaining datasets, including dynamic sequences, are then incorporated for the second stage of update operator finetuning and subsequent motion mask detector training.

\subsection{Model Architecture}
Our update operator is extended from DROID-SLAM~\cite{teed2021droid}. The full architecture is illustrated in \figref{fig:update}. Given a pair of input images ($\mathbf{I}_i$, $\mathbf{I}_j$), we first extract 3D-aware features from the pre-trained MASt3R Encoder. These high-dimensional features (1024 channels) are produced at $\frac{1}{16}$ resolution relative to the original image. We then utilize two separate encoders, the Context Feature Extractor and the Flow Feature Extractor, to process the MASt3R output. The architecture shared by these encoders is shown in \figref{fig:feat_extractor}. This stage lifts the feature resolution to the required $\frac{1}{8}$ scale. The Flow Feature Extractor output is subsequently used to compute the 4D correlation volume between the image features. The key component of the update operator is the ConvGRU layer. It takes the context feature, flow feature correlation, current hidden state, and the optical flow from the last iteration as input, and outputs an updated hidden state, which is further used to extract intermediate variables for differentiable BA, such as the updated optical flow and the confidence weight.

\section{Additional Results}
\label{sec:add_results}

\begin{table*}[t]
\centering
\footnotesize
\setlength{\tabcolsep}{4.5pt}
\begin{tabular}{lrrrrrrrrr}
\toprule
Method & \texttt{360} & \texttt{desk} & \texttt{desk2} & \texttt{floor} & \texttt{plant} & \texttt{room} & \texttt{rpy} & \texttt{teddy} & \texttt{xyz}\\
\midrule
\multicolumn{10}{l}{\cellcolor[HTML]{EEEEEE}{\textit{Keyframe Poses}}} \\
MASt3R-SLAM~\cite{murai2025mast3r} & \textbf{0.049} & \textbf{0.016} & \textbf{0.024} & 0.025 & 0.020 & 0.061 & 0.027 & 0.041 & \textbf{0.009}\\
VGGT~\cite{wang2025vggt} & \textbf{0.049} & 0.023 & 0.029 & 0.084 & 0.024 & 0.095 & 0.025 & 0.033 & 0.012\\
$\pi^{3}$~\cite{wang2025pi} & 0.065 & 0.018 & 0.028 & 0.085 & 0.026 & 0.061 & 0.024 & \textbf{0.028} & 0.011\\

\hdashline
\noalign{\vskip 1pt}
\multicolumn{10}{l}{\cellcolor[HTML]{EEEEEE}{\textit{Full Trajectory}}} \\
DROID-SLAM~\cite{teed2021droid} & 0.111 & 0.018 & 0.042 & 0.021 &\textbf{ 0.016} & 0.049 & 0.026 & 0.048 & 0.012\\
WildGS-SLAM~\cite{zheng2025wildgs} & 0.069 & 0.018 & 0.025 & 0.023 & 0.044 & 0.061 & 0.024 & 0.058 & \textbf{0.009}\\
MegaSaM~\cite{li2024megasam} & 0.055 & 0.019 & 0.040 & 0.278 & 0.038 & 0.080 & 0.024 & 0.043 & 0.013\\
ViPE~\cite{huang2025vipe} & 0.067& 0.024 & 0.035 & 0.320 & 0.025 & 0.048 & 0.020 & 0.038 & 0.011\\
{\textbf{\project{} (\ours)}} & 0.057 & \textbf{0.016} & \textbf{0.024} & \textbf{0.019} & 0.018 & 0.039 & \textbf{0.018} & 0.035 & \textbf{0.009} \\

\bottomrule
\end{tabular}

\caption{\textbf{Tracking Performance on TUM RGB-D (static) Dataset~\cite{sturm2012benchmark}} (ATE RMSE $\downarrow$ [m]). 
} 
\label{tab:tum_static}
\end{table*}

\begin{table*}[t]
\centering
\footnotesize
\setlength{\tabcolsep}{4.5pt}
\begin{tabular}{lrrrrrrr}
\toprule
Method & \texttt{chess} & \texttt{fire} & \texttt{heads} & \texttt{office} & \texttt{pumpkin} & \texttt{kitchen} & \texttt{stairs} \\
\midrule
\multicolumn{8}{l}{\cellcolor[HTML]{EEEEEE}{\textit{Keyframe Poses}}} \\
MASt3R-SLAM~\cite{murai2025mast3r} & 0.053 & 0.025 & \textbf{0.015} & 0.097 & \textbf{0.088} & 0.041 & \textbf{0.011} \\
VGGT~\cite{wang2025vggt} & 0.039 & 0.027 & 0.017 & 0.098 & 0.128 & 0.059 & 0.036 \\
$\pi^{3}$~\cite{wang2025pi} & \textbf{0.034} & 0.034 & 0.017 & 0.080 & 0.118 & \textbf{0.037} & 0.047 \\

\hdashline
\noalign{\vskip 1pt}
\multicolumn{8}{l}{\cellcolor[HTML]{EEEEEE}{\textit{Full Trajectory}}} \\
DROID-SLAM~\cite{teed2021droid} & 0.036 & 0.027 & 0.025 & 0.066 & 0.127 & 0.040 & 0.026 \\
WildGS-SLAM~\cite{zheng2025wildgs} & 0.037 & 0.032 & 0.027 & 0.054 & 0.154 & \textbf{0.037} & 0.019 \\
MegaSaM~\cite{li2024megasam} & 0.040 & 0.040 & 0.024 & 0.069 & 0.147 & 0.045 & 0.025 \\
ViPE~\cite{huang2025vipe} & 0.035 & \textbf{0.027} & 0.023 & 0.070 & 0.131 & 0.039 & 0.024\\
{\textbf{\project{} (\ours)}} & 0.038 & 0.029 & 0.024 & \textbf{0.053} & 0.142 & \textbf{0.037} & 0.023  \\

\bottomrule
\end{tabular}

\caption{\textbf{Tracking Performance on 7-Scenes Dataset~\cite{shotton2013scene}} (ATE RMSE $\downarrow$ [m]). 
} 
\label{tab:7scenes}
\end{table*}

\paragraph{Full Tracking Results on the static Dataset} In the main paper, we summarize the average ATE for the TUM RGB-D (static)~\cite{sturm2012benchmark} and 7-Scenes~\cite{shotton2013scene} datasets. Here, we present the results of full sequences in \tabref{tab:tum_static} (TUM RGB-D) and \tabref{tab:7scenes} (7-Scenes). On the TUM RGB-D sequences, our method shows superior performance compared to the baselines, achieving the best result on 5 out of 9 sequences. Notably, compared to recent dynamic methods like MegaSaM (ATE: $0.278 \text{ m}$) and ViPE (ATE: $0.320 \text{ m}$), \project{} exhibits higher stability on the challenging \texttt{floor} sequence (ATE: $0.020 \text{ m}$). In this sequence, some temporally close frames contain large relative motion. This rapid change in viewpoint reduces feature overlap and induces large pixel displacements. Our resilience on this sequence stems from the broad variety of camera motion types included in our training data and the incorporation of loop closure. Furthermore, on the 7-Scenes dataset, where most methods exhibit only marginal performance differences, \project{} maintains competitive accuracy across all sequences. The consistently strong behavior across these benchmarks affirms the robustness of our method in static environments.

\vspace{8pt}
\begin{table}[t]
    \centering
    \vspace{-8pt}
    \footnotesize
    \newcommand{\sz}{0.195}
    \newcommand{\sza}{0.1729} 
    \setlength{\tabcolsep}{3pt}
    \resizebox{0.95\linewidth}{!}{
    \begin{tabular}{lccccccc}
    \toprule
    Method & \textit{0000}& \textit{0059}& \textit{0106}& \textit{0169}& \textit{0181}& \textit{0207} & Avg. \\
    \midrule
    DROID-SLAM~\cite{teed2021droid} & \textbf{5.48} & 9.00 & 6.76 & \textbf{7.89} & 7.41 & 7.97 & 7.42\\
    WildGS-SLAM~\cite{zheng2025wildgs} & 5.74 &8.15 & 9.15 &8.31 &7.17 &7.31 & 7.64\\
    MegaSaM~\cite{li2024megasam} & \textit{OOM} & 8.56 & 37.35 & 10.16 & 8.35 & 8.54 & -\\
    \textbf{WildPose (Ours)} &6.11 & \textbf{8.07} & \textbf{6.75} & 8.01 & \textbf{6.56}&\textbf{ 7.36} &\textbf{7.14} \\
    \bottomrule
    \end{tabular}}
    \caption{\textbf{Tracking Performance on ScanNet Dataset~\cite{dai2017scannet}} (ATE RMSE $\downarrow$ [m]). 
} 
    \label{tab:scannet}
\end{table}
\paragraph{Addtional Evaluation Results on the ScanNet dataset~\cite{dai2017scannet}}
To evaluate our method's robustness to motion blur and its generalization across diverse static indoor environments, we conduct additional experiments on the ScanNet dataset. The tracking results are reported in \tabref{tab:scannet}. DROID-SLAM results are sourced from GO-SLAM and Nicer-SLAM [56]. WildPose still outperforms other baselines in this static benchmark.


\begin{table}[t]
    \centering
    \vspace{2pt}
    \footnotesize
    \setlength{\tabcolsep}{3pt}
    \resizebox{0.95\linewidth}{!}{
    \begin{tabular}{lccc}
    \toprule
    Method
     & ATE [cm] $\downarrow$ & Avg. fps $\uparrow$ & {Peak GPU Use [GiB] $\downarrow$} \\
    \midrule
    DROID-SLAM~\cite{teed2021droid} & 16.17 & \textbf{11.27} & \textbf{7.52} \\
    WildGS-SLAM~\cite{zheng2025wildgs} & 0.46 & 0.49 & 14.65 \\
    MegaSaM~\cite{li2024megasam} & 2.40  & 1.86 & 21.87\\
    Vipe~\cite{huang2025vipe} & 2.59 & 6.44 & 13.45 \\
    \textbf{WildPose (Ours)} & \textbf{0.39} & 2.98 & 18.62 \\
    \bottomrule
    \end{tabular}}
    \caption{\textbf{Run time and memory usage on Wild-SLAM~\cite{zheng2025wildgs}.} We compute FPS by dividing the total number of frames by the total running time. The experiments are conducted on an RTX 4090 GPU.}
    \label{tab:time_analysis}
    \vspace{-8pt}
\end{table}

\paragraph{Runtime Analysis} We compare the average processing frame rate (FPS) and the peak GPU usage of our method against the baselines that estimate poses of the full trajectory, with detailed results presented in \tabref{tab:time_analysis}. For a fair comparison, MegaSaM’s FPS includes depth preprocessing but excludes video depth optimization. While lightweight systems (DROID-SLAM, Vipe) are faster, they struggle with dynamic scenes. Compared to more complex dynamic frameworks like MegaSaM and WildGS-SLAM, WildPose offers superior tracking accuracy (lowest ATE) with a higher FPS. Our peak memory usage stems from foundation models (Moge2, MASt3R) but is mitigatable via preprocessing, similar to MegaSaM, or using a distilled model.

\section{Limitations}
\label{sec:limitation}

\begin{figure}[t]
    \centering
    
    \begin{subfigure}[b]{0.48\columnwidth}
        \centering
        \includegraphics[width=\linewidth]{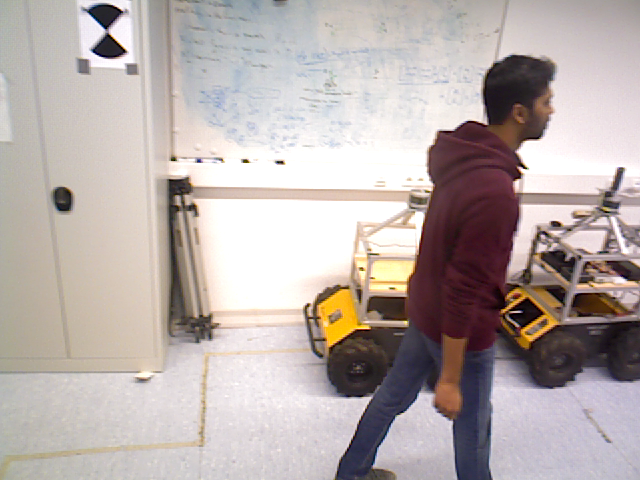}
    \end{subfigure}
    \hfill 
    \begin{subfigure}[b]{0.48\columnwidth}
        \centering
        \includegraphics[width=\linewidth]{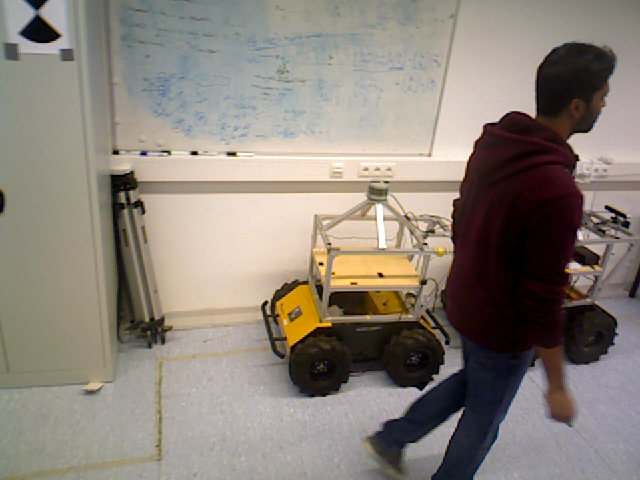}
    \end{subfigure}

    \vspace{3mm} 
    
    \begin{subfigure}[b]{0.48\columnwidth}
        \centering
        \includegraphics[width=\linewidth]{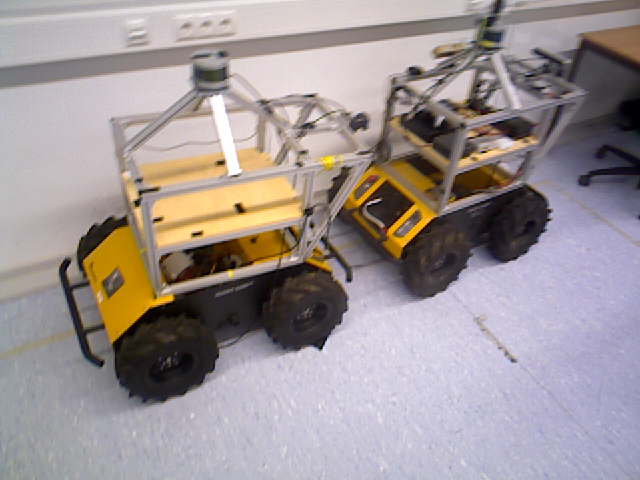}
    \end{subfigure}
    \hfill 
    \begin{subfigure}[b]{0.48\columnwidth}
        \centering
        \includegraphics[width=\linewidth]{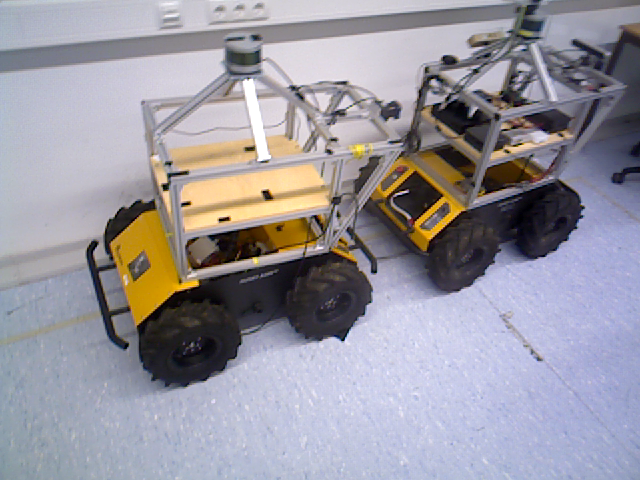}
    \end{subfigure}
    
    \caption{\textbf{Limitations.} We visualize sampled images from Bonn RGB-D Dynamic Dataset~\cite{palazzolo2019iros} (\texttt{Person} sequence). The dataset has inconsistent exposure, which is challenging to our approach.}
    \label{fig:bonn_limit}
\end{figure}

\project{}’s learnable modules are trained exclusively on synthetic data. Although our curriculum is diverse, a domain gap to real-world scenarios inevitably exists. This gap is evident in sequences with unobserved phenomena, such as the significant photometric variations in the Bonn RGB-D Dynamic Dataset (\figref{fig:bonn_limit}). Our model, lacking explicit training on such exposure changes, achieves slightly worse accuracy on this dataset, particularly on the \texttt{Person} and \texttt{Person 2} sequences. Furthermore, our framework assumes known and fixed camera intrinsics, limiting its applicability to uncalibrated videos. Addressing these gaps through domain adaptation and online intrinsic refinement remains an important avenue for future work.


\end{document}